\documentclass{article} 
\usepackage{iclr2026_conference,times}

\usepackage{amsmath,amsfonts,bm}

\def\eqref#1{equation~\ref{#1}}

\def\1{\bm{1}}

\DeclareMathAlphabet{\mathsfit}{\encodingdefault}{\sfdefault}{m}{sl}
\SetMathAlphabet{\mathsfit}{bold}{\encodingdefault}{\sfdefault}{bx}{n}

\usepackage{hyperref}
\usepackage{url}
\usepackage{graphicx}
\usepackage{booktabs}
\usepackage[T1]{fontenc}    

\usepackage{subcaption}

\usepackage{amsmath}
\usepackage{amssymb}
\usepackage{mathtools}
\usepackage{amsthm}
\usepackage{threeparttable}
\usepackage{placeins}
\usepackage{multirow}

\usepackage[capitalize,noabbrev]{cleveref}
\newtheorem{theorem}{Proposition}

\title{Rewarding Doubt: A Reinforcement Learning Approach to Calibrated Confidence Expression of Large Language Models}

\author{David Bani-Harouni\textsuperscript{1,2}\thanks{Equal contribution} \And Chantal Pellegrini\textsuperscript{1,2}\footnotemark[\value{footnote}] \And Paul Stangel\textsuperscript{1} \And Ege Özsoy\textsuperscript{1,2} \And Kamilia Zaripova\textsuperscript{1,2} \And Nassir Navab\textsuperscript{1,2} \And Matthias Keicher\textsuperscript{1,2} \AND\\[-1em]
\textsuperscript{1}Technical University of Munich \\
\textsuperscript{2}Munich Center for Machine Learning\\
\texttt{\{david.bani-harouni,chantal.pellegrini\}@tum.de}}

\iclrfinalcopy 
\begin{document}

\maketitle

\begin{abstract}
A safe and trustworthy use of Large Language Models (LLMs) requires an accurate expression of confidence in their answers. We propose a novel Reinforcement Learning approach that allows to directly fine-tune LLMs to express calibrated confidence estimates alongside their answers to factual questions. Our method optimizes a reward based on the logarithmic scoring rule, explicitly penalizing both over- and under-confidence. This encourages the model to align its confidence estimates with the actual predictive accuracy. The optimal policy under our reward design would result in perfectly calibrated confidence expressions. Unlike prior approaches that decouple confidence estimation from response generation, our method integrates confidence calibration seamlessly into the generative process of the LLM. Empirically, we demonstrate that models trained with our approach exhibit substantially improved calibration and generalize to unseen tasks without further fine-tuning, suggesting the emergence of general confidence awareness. Our code is available at \href{https://github.com/pasta99/RewardingDoubt}{\texttt{https://github.com/pasta99/RewardingDoubt}}.
\end{abstract}

\section{Introduction}
\label{sec:intro}
In human intelligence and inter-human interaction, the ability to understand our own uncertainty and communicate our doubts to others is fundamental for effective decision-making, collaboration, and learning \citep{cosmides1996humans,xiong2024llmsexpressuncertaintyempirical}. Similarly, for Large Language Models (LLMs) to be safely used in real-world applications, especially when humans and AI systems work together, they must not only generate accurate information but also communicate their confidence in that information. While LLMs have demonstrated impressive capabilities in natural language understanding, question answering and text summarization \citep{touvron2023llama,vicuna2023,openai2023gpt4}, LLMs still face significant limitations, such as their tendency to generate inaccurate information, often referred to as hallucinations \citep{hadi2023survey}. This raises concerns about their reliability, particularly in real-world applications where trustworthiness is essential. Especially in high-stakes environments such as medical diagnosis, where LLMs are starting to become support tools for professionals \citep{moor2023foundation,pellegrini2025radialog,tu2024towards,bani2024magda}, overconfident predictions including factual errors or hallucinations could have serious consequences for patient health. Also, in customer service or legal consultation \citep{shi2024chopschatcustomerprofile,sun2024lawluomultiagentcollaborativeframework}, LLMs need to express uncertainty and defer complex queries to human representatives when unsure to avoid misinformed decisions. Reliable confidence estimation and expression would enable these systems to flag uncertain outputs for human review, ensuring that crucial decisions are not made based on uncertain LLM outputs. To allow risk estimation while using LLM-generated output, model confidence should be calibrated, meaning that the expressed numerical confidence should be equal to the probability of the model's answer being correct.

\begin{figure}[tb]
    \centering
    \includegraphics[width=0.8\linewidth]{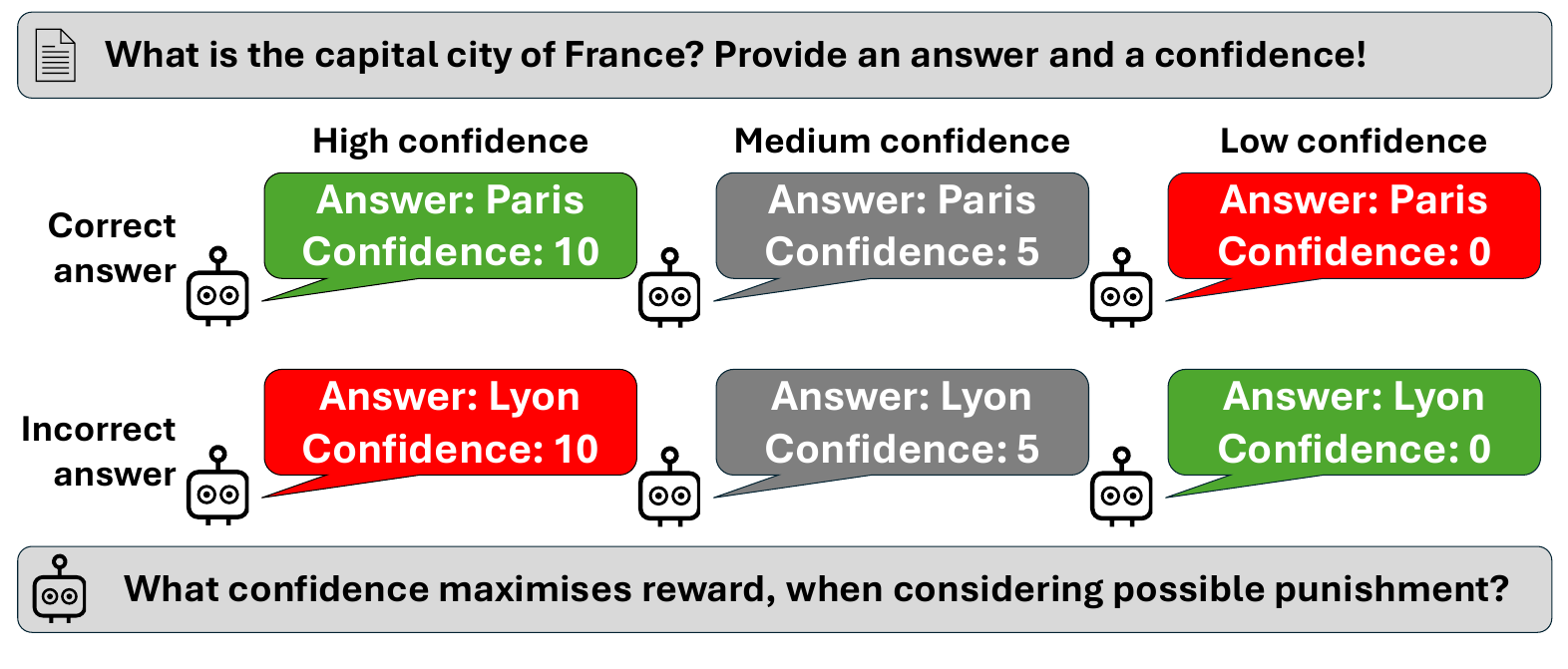}
    \caption{Illustration of our reward design: The model is rewarded for a high confidence if it is correct and punished if it is incorrect. To maximize the reward, the model needs to learn when to predict a higher or lower confidence, considering a possible higher punishment. Our reward function is designed so that the best reward is given when the confidence and the epistemic probability of being correct are the same, thus incentivizing the expression of calibrated confidences. }
    \label{fig:schaubild}
\end{figure}

Many previous methods for confidence estimation lack in calibration performance as they do not train the model and instead infer the confidence from the internal state in a zero-shot setup \citep{huang2023lookleapexploratorystudy, kuhn2023semantic, duan2024shifting}. Additionally, this does not give models an inherent awareness of confidence. Other trained methods in this area decouple the uncertainty estimation from the text generation process \citep{azaria2023internalstatellmknows, kapoor-etal-2024-calibration}. This approach optimizes for calibrated confidence estimation but does not enable the uncertainty-awareness and expression in the model itself. 

Targeting these limitations, we propose a novel reinforcement-learning (RL) approach for teaching LLMs to express their calibrated confidence, encouraging a granular, accurate estimation of the confidence level in the training objective. For this, we model confidence estimation as a betting game: a high-confidence answer would warrant a larger bet, reflecting a strong belief in its correctness, while a lower confidence score would suggest caution. Central to our method is a reward function based on the logarithmic scoring rule, a strictly proper scoring rule. We are the first to optimize this function through reinforcement-learning-based policy optimization, leveraging its calibration properties for directly and seamlessly training confidence calibration in LLM generations. This reward function captures the fundamental risk-reward tradeoff in probabilistic decision-making, as illustrated in \cref{fig:schaubild}. It increases the reward when a correct answer is given with high confidence, simulating the higher potential return of big bets. Conversely, it penalizes incorrect answers more when they are made with high confidence, discouraging overconfidence. This ensures that both uncertainty and confidence are appropriately factored into the reward system. As a proper scoring rule, optimizing the reward function trains the model to align its predicted confidence with the accuracy of its output, encouraging granular and calibrated confidence scoring. A calibrated confidence estimation will provably result in the highest reward during training. This not only improves the trustworthiness of LLMs in collaborative human-AI scenarios but also helps users better assess when AI tools should be trusted, double-checked, or deferred to human expertise.

\section{Related Works}
\subsection{Confidence Estimation in LLMs}
Confidence estimation and calibration have a long history in machine learning and natural language processing \citep{wang2024calibrationdeeplearningsurvey}. With the rise of LLMs, research has focused on adapting and extending these ideas to modern architectures. Broadly, methods fall into black-box and white-box approaches \citep{geng2024surveyconfidenceestimationcalibration}.

\paragraph{Black-box methods} Black-box methods rely only on model outputs. Linguistic prompting methods ask the model to verbalize its confidence, sometimes aided by chain-of-thought reasoning \citep{xiong2024llmsexpressuncertaintyempirical,wei2022chain}. Consistency-based approaches estimate confidence by measuring agreement across multiple generations, with high variance indicating uncertainty \citep{manakul2023selfcheckgpt,wang2022self}. Recently, \citet{zhou2025steerconf} proposed SteerConf, which does multiple inference passes where the LLM is prompted to use different levels of caution in its confidence expression. The resulting verbalized confidences are aggregated based on confidence and answer consistency to an overall confidence prediction. Black-box methods are valuable for their simplicity, ease-of-use and universality, however generally lack behind white-box methods in their calibration performance.

\paragraph{White-box methods} White-box methods exploit internal model states. Logit-based techniques estimate confidence from token probabilities or entropy \citep{huang2023lookleapexploratorystudy,kuhn2023semantic,duan2024shifting}, assuming that high probability tokens correspond to high confidence predictions. Self-evaluation methods let the model judge the truth of its own answers \citep{kadavath2022languagemodelsmostlyknow}. They prompt the model to provide an answer followed by a judgment whether its own answer is "true" or "false". They then compare the probability of the "true" or "false" token to calculate a confidence estimation. External probing approaches train classifiers on hidden states to predict correctness \citep{azaria2023internalstatellmknows}. While some of these methods achieve good confidence estimation results, they do not teach the model to express clear confidence values itself but depend on some auxiliary estimation mechanism.

\subsection{Finetuned Confidence Expression}
A growing line of work integrates confidence estimation into instruction tuning. These methods typically follow a two-step paradigm: First, they estimate model confidence using various methods, e.g., self-consistency \citep{cheng2024can,yang2024alignment,han2024enhancing}, token probabilities \citep{chen2024teaching}, trained probes \citep{mielke2022reducing}, empirical accuracy \citep{zhang2024r,lin2022teaching,ulmer2024calibrating}, or topic unfamiliarity \citep{wan2024knowledge, kang2024unfamiliar}. Second, they construct finetuning datasets that either replace uncertain answers with refusals \citep{zhang2024r,cheng2024can,yang2024alignment,wan2024knowledge} or append the estimated uncertainty as an additional supervised signal \citep{han2024enhancing,chen2024teaching,mielke2022reducing,lin2022teaching,ulmer2024calibrating}.

The key limitation of this approach is that the model’s expressed confidence is bounded by the quality of the constructed ground-truth estimates. Additionally, while the underlying confidence estimation method might optimize for perfect calibration (e.g. in the case of the trained probe), this theoretical guarantee is lost when performing supervised finetuning on these constructed ground truths to reproduce these scores.

\subsection{Reinforcement Learning for Confidence Expression}
Reinforcement Learning from Human Feedback (RLHF) has proven effective for aligning LLMs with human preferences \citep{ouyang2022training}, and has also been explored for agentic interaction in textual environments \citep{zhou2023dialogue,carta2023grounding}. Only recently have researchers begun applying RL directly to confidence estimation. \citet{tao2024trust} adapt RLHF by designing rewards that align verbalized confidence with preference ratings, but this requires human-annotated preference data and does not address factual calibration. \citet{Leng2024TamingOI} identify that standard reward models in RLHF are biased toward high verbalized confidence, rating answers with high confidence expressions with a high reward. To counteract this, they introduce two reward model training paradigms, PPO-M and PPO-C, which fine-tune the reward model to reward answers where correctness and confidence expression are aligned. \citet{xu2024rejectionimprovesreliabilitytraining} propose RL from Knowledge Feedback (RLKF) to encourage refusals outside the model’s knowledge scope, reducing hallucinations but without quantifying confidence. \citet{stengel2024lacie} propose LACIE, a DPO-based approach that simulates an interaction between a speaker and a listener model, rewarding accurate and honest confidence expression by aligning it with the listener’s interpretation of confidence cues rather than with fact-based numerical calibration.

In contrast to previous works, our method directly optimizes for factual calibration using a theoretically grounded, proper scoring rule as the reward signal, enabling the model to develop intrinsic uncertainty awareness without requiring external preference models, knowledge supervision, or post-hoc calibration techniques, while at the same time seamlessly integrating calibrated confidence expression into the LLMs response generation.

\section{Rewarding Doubt}
\label{sec:method}
\begin{figure}[tb]
    \centering
    \includegraphics[width=\linewidth]{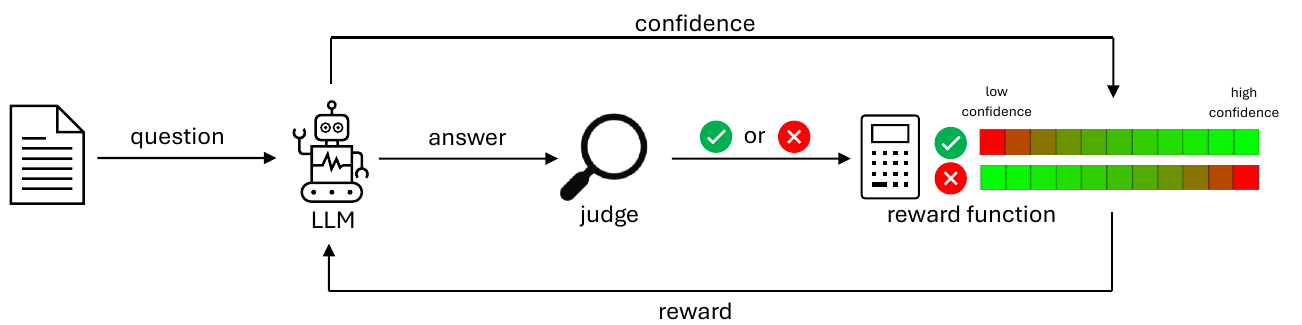}
    \caption{Overview of our reinforcement learning framework: The LLM is prompted to answer a question and provide the confidence in this answer. The answer is checked for correctness by a judge function and the reward is calculated based on the correctness and the confidence. Correct answers with high confidences are rewarded highly, but also penalized heavily when incorrect. }
    \label{fig:framework}
\end{figure}

We propose a novel reinforcement-learning approach, that improves an LLM's ability to verbalize an accurate numerical confidence in a previously generated answer. The LLM functions as an agent in a simulated environment as shown in \cref{fig:framework}, that poses challenging question-answering scenarios. It is prompted with task queries such as factual questions and asked to predict both an answer to the query as well as a confidence score. Based on the correctness of the answer, and the expressed confidence, we reward the model, incentivizing it to express a calibrated confidence.

Formally, let the model be provided with a textual question or request $q$, resulting in an answer-confidence pair $(a,\hat{p})$ as response, where $a$ is a textual answer with binary correctness value, and $0 \leq \hat{p} \leq 1$ is a numerical confidence score representing the subjective probability the model assigns to answer $a$ being correct. We train this subjective probability assessment to align with the true epistemic probability $p^*$, which represents the actual likelihood of correctness given the model’s internal knowledge state. If $\hat{p}$ and $p^*$ are aligned the model is perfectly calibrated, meaning the probability of correctness $P(j(a)=1)$ always equals the expressed confidence:
\[P(j(a)=1 \mid \hat{p} = x) = x, \quad \forall x \in [0,1],\]
where $j(\cdot)$ is a correctness judging function that is 1 if answer $a$ is correct, and 0 otherwise.

The true epistemic probability $p^*$ is not directly observable, thus supervised learning of calibration is only possible by constructing an artificial ground truth to approximate $p^*$. Instead, we model this task as a Markov Decision Process (MDP) defined by the tuple $(\mathcal{S}, \mathcal{A}, \mathcal{T}, R)$, where the model learns to generate calibrated confidence scores through reinforcement learning. The MDP is defined by the following components:
\begin{itemize}
    \item \textbf{State space} ($\mathcal{S}$): A state $s_t \in \mathcal{S}$ consists of a natural language question $q$, the model’s predicted answer $a$, and the partial sequence of confidence tokens predicted so far, if any. That is, $s_t = (q, a, c_{1:t-1})$, where $c_{1:t-1}$ represents the previously generated confidence score tokens.
	\item \textbf{Action space} ($\mathcal{A}$): The action space consists of selecting the next token $c_t$ in the confidence estimation process from the LLM vocabulary, including numerical tokens (e.g., representing percentages or probability values) and a special end-of-sequence token that finalizes the prediction.
	\item \textbf{Transition function} ($\mathcal{T}(s_{t+1} \mid s_t, a_t)$): The environment transitions deterministically based on the language model’s autoregressive token generation process. Given a state $s_t = (q, a, c_{1:t-1})$ and an action $c_t$, the next state is defined as $s_{t+1} = (q, a, c_{1:t})$. Once the end-of-sequence token is generated, the episode terminates.
	\item \textbf{Reward function} ($R$): The reward $R(a, c, j)$ is computed based on the final confidence score sequence $c = (c_1, \dots, c_T)$ and the correctness of the answer $j(a)$.
\end{itemize}

To promote accurate confidence estimation, the model's expected reward must fulfill the requirement of being maximized when $\hat{p} = p^*$, i.e. when the predicted confidence aligns with the probability of correctness. The model should receive a high reward when it correctly predicts an outcome with high confidence or when it incorrectly predicts an outcome with low confidence. Conversely, the reward should be low when incorrect predictions are provided with high confidence or correct predictions are provided with low confidence. This approach incentivizes the model to express high confidence only in cases where certainty is warranted while expressing doubt in ambiguous situations. By penalizing both overconfidence and underconfidence, the model is encouraged to calibrate its confidence accurately, effectively balancing the trade-off between reward maximization and penalty avoidance. Note, that through this design our method focuses exclusively on improving calibration while keeping task performance stable.

We design our reward as a logarithmic scoring function:
\begin{equation}
\label{eq:rewardfunction}
    R(a, \hat{p}, j) = \begin{cases}
    log(\hat{p}), & \text{if $j(a)$ = 1 (correct)}\\
    log(1 - \hat{p}), & \text{\text{if $j(a)$ = 0 (incorrect)}}
    \end{cases}
\end{equation}

This function fulfills the requirement described above as we show in the following proposition:

\begin{theorem}[Optimality implies Calibration]
\label{prop_opt}
    The expected reward $\mathbb{E}[R(a, \hat{p}, j)]$ is maximized for each sample when $\hat{p} = p^*$ and the optimal policy under the reward design is thus perfectly calibrated.
\end{theorem} 
The proof of Proposition \ref{prop_opt} is analogous to the proof that the logarithmic scoring rule is a proper scoring rule. We provide it in full in \cref{sec:proofs} and discuss the influence of the clipping on the optimality of the reward function.

\begin{figure}
    \centering
    \includegraphics[width=0.7\linewidth]{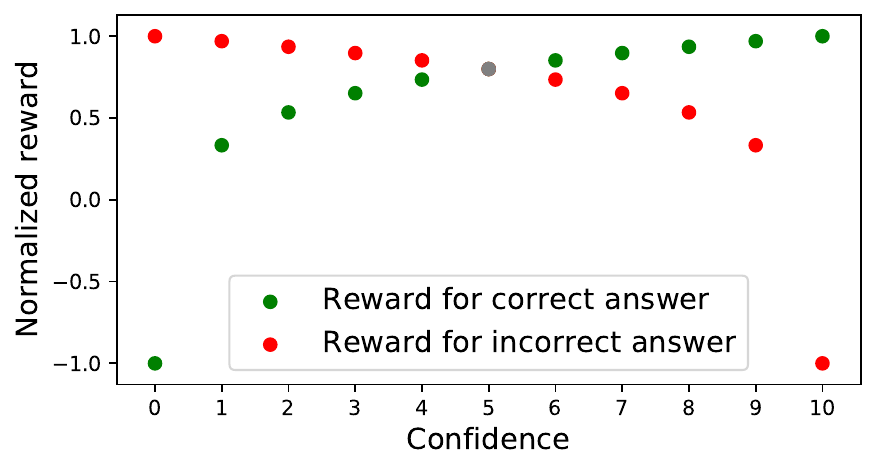}
    \caption{The rewards for each confidence value for correct and incorrect answers. The closer the confidence is to ten or zero, respectively, the higher is the reward. At the same time, the possible punishment increases to a greater extent. The model has to learn when the trade-off between those two possibilities is worthwhile.}
    \label{fig:rewardfunction}
\end{figure}

Since the logarithm of zero is undefined, we introduce a small positive constant $\epsilon$ as clipping value for numerical stability. Concretely, we clip the lower and upper limit of the confidence $\hat{p}$ to $\epsilon$ and $1-\epsilon$, respectively. The clipped reward function is provided in \cref{app:clipped_reward}. 
The normalized and clipped reward for correct and incorrect answers for each confidence is visualized in \cref{fig:rewardfunction}. 

\section{Experimental Setup}
\label{sec:exp}
We evaluate our method in both Single-Answer and Multiple-Answer settings. We prompt the model to provide a confidence for each answer as an integer between 0 and 10, which we normalize for the reward calculation. A confidence of zero is defined as the model being certain that the answer is incorrect, while ten is defined as the model being certain the answer is correct. We normalize the reward function to the range of $[-1, 1]$.

In the Single-Answer setting we train the model on the TriviaQA dataset \citep{JoshiTriviaQA2017}, which contains question-answer-evidence triplets, from which we only use the questions and answers. For generalization experiments, we evaluate our method on CommonsenseQA \citep{talmor-etal-2019-commonsenseqa} and MedQA \citep{jin2020diseasedoespatienthave}, which are multiple-choice question datasets in the commonsense and medical domain, respectively. For the Multiple-Answer setting, we train on the QAMPARI dataset \citep{amouyal2023qampariopendomainquestionanswering}, which contains questions with multiple-answers as well as evidence, again only using the questions and answers.

In the Single-Answer setting we compare our approach on the TriviaQA dataset against the following methods: Chain-of-Thought \citep{xiong2024llmsexpressuncertaintyempirical}, Top-K \citep{tian2023just}, Surrogate Token \citep{kadavath2022languagemodelsmostlyknow}, Sequence Probability \citep{huang2023lookleapexploratorystudy} and Self-Consistency \citep{wang2022self} as zero-shot methods, LACIE \citep{stengel2024lacie}, which uses DPO for optimizing confidence expression and Trained Probe \citep{azaria2023internalstatellmknows}, which employs supervised training of an external probe for estimation model. We also compare to the non-finetuned base model in a zero-shot manner, using the same prompt as our Rewarding Doubt method and refer to this setup as Verbalize.
In the Multiple-Answer setting we compare to Trained Probe and Sequence probability, as those methods are the best performing zero-shot and trained baselines in the Single-Answer setting. LACIE does not report results for this dataset, thus we can only compare on TriviaQA.

We report our results using the Expected Calibration Error (ECE) and the Area Under the Receiver Operating Characteristic Curve (AUROC) metric. 
Additionally, we visualize the calibration with calibration curves, where a well-calibrated model lies close to the 45° line and large deviations show a high miscalibration.

\paragraph{Response Generation}
To calibrate and reward the model only on the confidences and not the answers we separate generation in two steps during training: Answer and confidence generation. Answers are generated first and afterwards treated as fixed inputs alongside the question, while the confidence is generated in a separate generation step and considered as sole target for optimization. Like this, we ensure that answer generation is disentangled from the optimization process, ensuring the answer correctness is not affected by our confidence calibration training.

\paragraph{Correctness Assessment}
For the multiple-choice datasets MedQA and CommonsenseQA, we evaluate correctness using the exact string matching between the model's response and the ground truth answer. For the TriviaQA and QAMPARI datasets, we use the F1 score of word overlap to measure the similarity between the model's response and the ground truth candidates. The F1 score is calculated for each candidate and the maximum score is considered the final score. We consider an answer as correct if its score exceeds a threshold of 0.5.

\paragraph{Implementation Details}
\label{impl}
We optimize the reward function using the Proximal Policy Optimization (PPO) algorithm \citep{schulman2017proximal}. Unless stated otherwise, we use Meta-Llama-3-8B-Instruct \citep{grattafiori2024llama3herdmodels} as base model for our experiments. We employ the 4-bit quantized performance-optimized model version by Unsloth AI \citep{unsloth} and apply LoRA fine-tuning \citep{hu2022lora}. For the Single-Answer setting we train the model for two epochs with a learning rate of 1e-5. For the Multiple-Answer setting, due to the size of the training dataset and the fact that each question yields multiple facts, the model is trained for a limited amount of 24,000 steps with a batchsize of eight and a learning rate of 1e-5 and multiply the reward with 5 to increase its spread. All models are trained on one Nvidia A40 with each training run taking seven days. On average the model generated approximately 3.4 answers per fact. If the model fails to generate an answer in the specified format, it is penalized with an out-of-format reward of -3. Detailed implementation choices for the baselines are provided in \cref{app:baselines}.

\section{Results and Discussion}
\label{sec:results}

This section presents and discusses the key findings of our experiments for both Single and Multiple-Answer tasks and the generalization to out-of-domain datasets.

\begin{table}[htb]
\caption{Comparison of methods on the TriviaQA dataset in the Single-Answer setting with 95\% CIs in brackets. * Results are from the original paper \citep{stengel2024lacie} and include standard error.}
\label{tab:comparison_methods_triviaqa}
\centering
\begin{tabular}{lccc}
\toprule
\textbf{Method} & \textbf{ECE} ($\downarrow$) & \textbf{AUROC} ($\uparrow$) & \textbf{Accuracy } ($\uparrow$)\\
\midrule
Verbalize & 0.3459 \scriptsize[0.3375,0.3543] & 0.5858  \scriptsize[0.5778,0.5936] & 0.6310  \scriptsize[0.6222,0.6397]\\
Chain-of-Thought & 0.3065 \scriptsize[0.2981,0.3157] & 0.6379 \scriptsize[0.6284,0.6475] & 0.6273 \scriptsize[0.6181,0.6363]\\
Top-K & 0.1611 \scriptsize[0.1529,0.1695] & 0.6673 \scriptsize[0.6580,0.6768] & 0.6023 \scriptsize[0.5936,0.6110]\\
Surrogate Token & 0.3686 \scriptsize[0.3595,0.3783] & 0.5923 \scriptsize[0.5818,0.6027] & 0.5933 \scriptsize[0.5844,0.6016]\\
Sequence Probability & 0.3156 \scriptsize[0.3074,0.3237] & 0.7804 \scriptsize[0.7725,0.7876] & 0.5955 \scriptsize[0.5864,0.6040]\\
Self-Consistency & 0.1134 \scriptsize[0.1066,0.1210] & 0.8213 \scriptsize[0.8129,0.8298] & 0.6224 \scriptsize[0.6131,0.6317]\\
PPO-M & 0.3262  \scriptsize[0.3173,0.3346] & 0.5274  \scriptsize[0.5227,0.5319] & 0.5749  \scriptsize[0.5662,0.5835]\\
PPO-C & 0.3607  \scriptsize[0.3524,0.3697] & 0.5439  \scriptsize[0.5384,0.5491] & 0.5258  \scriptsize[0.5164,0.5358]\\
LACIE* & 0.1200 \scriptsize$\pm$0.02 & 0.7200 \scriptsize$\pm$0.02 & n/a\\
Trained Probe & \textbf{0.0189} \scriptsize[0.0147,0.0275] & 0.8173 \scriptsize[0.8099,0.8250] & 0.5925 \scriptsize[0.5834,0.6017]\\
Rewarding Doubt (ours) & 0.0226 \scriptsize[0.0176,0.0302] & \textbf{0.8592} \scriptsize[0.8523,0.8664] & 0.6309 \scriptsize[0.6222,0.6399]\\
\bottomrule
\end{tabular}
\end{table}

To assess how well our approach improves calibration, we compare it against the zero-shot LLM baseline (Verbalize) and several established methods in both Single-Answer and Multiple-Answer question-answering tasks. Results for the Single-Answer setting on TriviaQA are presented in \cref{tab:comparison_methods_triviaqa}, and those for the Multiple-Answer setting on QAMPARI appear in \cref{tab:comparison_qampari}. Across both tasks, Rewarding Doubt substantially improves the model’s confidence calibration over zero-shot verbalization.

In the Single-Answer setting on TriviaQA, Rewarding Doubt achieves an ECE of 0.0226 and an AUROC of 0.8592, clearly outperforming all zero-shot baselines as well as LACIE, which is based on DPO-based optimization. The second fine-tuned method, Trained Probe, which relies on supervised fine-tuning, reports a slightly lower ECE (0.0189), both methods achieve near-perfect results. Further the AUROC of Rewarding Doubt is notably higher, suggesting that although both methods offer strong calibration, Rewarding Doubt better discriminates between correct and incorrect answers. In the Multiple-Answer setting on QAMPARI, Rewarding Doubt also outperforms baselines, achieving an ECE of 0.0816 and an AUROC of 0.6947. In comparison, Verbalize, Sequence Probability, and Trained Probe perform notably worse. Our findings support the claim by \citet{azaria2023internalstatellmknows} that a model’s internal state encodes information about the truthfulness of statements, which can serve as an indicator of uncertainty. However, without fine-tuning, the model struggles to utilize this internal information effectively. Our approach enables the model to make use of this correlation and translate it into an accurate expression  of the probability that a given answer is correct.

\begin{figure}[tb]
    \centering
    \begin{subfigure}[t]{0.48\textwidth}
        \includegraphics[width=\textwidth]{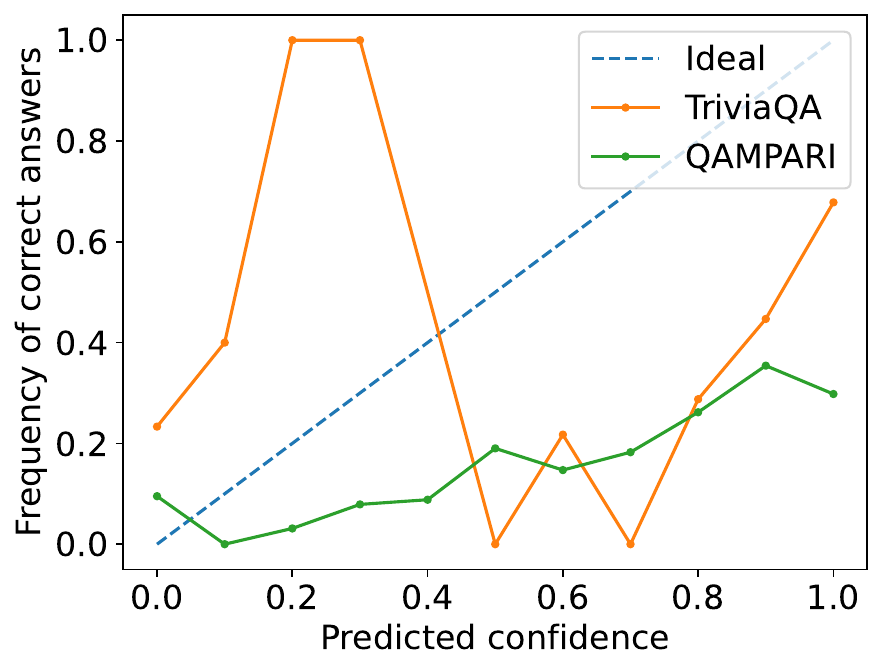}
    \caption{Base model}
    \label{fig:calibuntrained}
    \end{subfigure}
    \hfill
    \begin{subfigure}[t]{0.48\textwidth}
        \includegraphics[width=\textwidth]{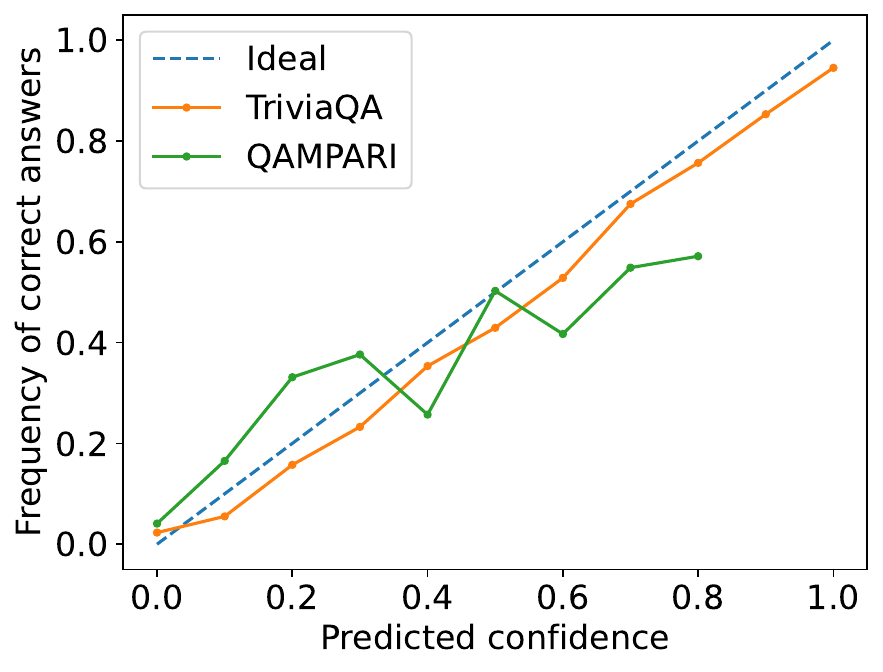}
        \caption{Fine-tuned by Rewarding Doubt}
        \label{fig:calibtrained}
    \end{subfigure}
    \caption{Calibration curves of the zero-shot base model (Verbalize) and the model fine-tuned by Rewarding Doubt.}
    \label{fig:calibs}
\end{figure}

\begin{figure}[tb]
    \centering
    \begin{subfigure}[t]{0.45\textwidth}
        \includegraphics[width=\textwidth]{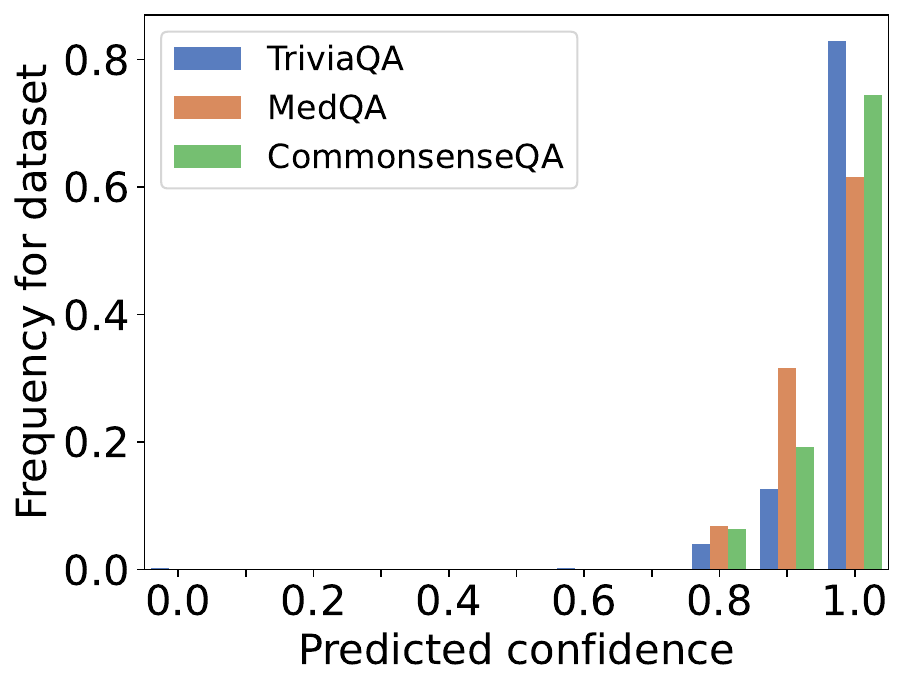}
    \caption{Base model}
    \label{fig:histogramuntrained}
    \end{subfigure}
    \hfill
    \begin{subfigure}[t]{0.45\textwidth}
        \includegraphics[width=\textwidth]{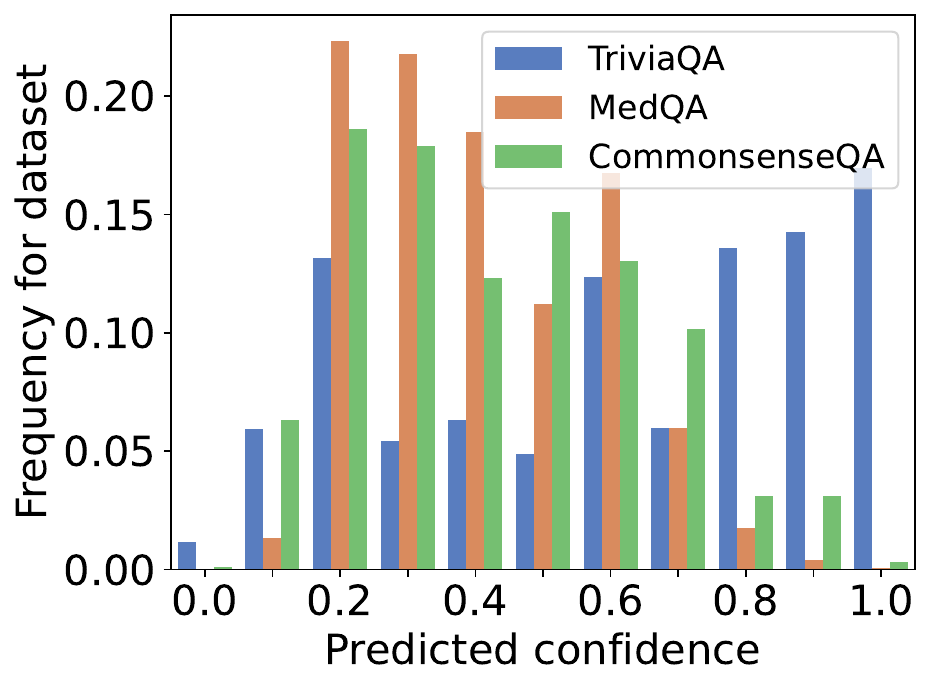}
        \caption{Fine-tuned by Rewarding Doubt}
        \label{fig:histogramtrained}
    \end{subfigure}
    \caption{Histograms of predicted confidences of the zero-shot base model (Verbalize) and the model fine-tuned on the TriviaQA dataset. }
    \label{fig:enter-label}
\end{figure}

The calibration curves in \cref{fig:calibs} further illustrate these improvements. For both TriviaQA and QAMPARI, the fine-tuned model's confidence much more closely aligns with the ideal 45° line than the zero-shot Verbalize baseline. Additionally, we observe a shift in the confidence distribution after fine-tuning. As shown in \cref{fig:enter-label}, in a zero-shot setting the LLM (Verbalize) predominantly assigns high confidence scores (8 or above), reflecting overconfidence, a pattern also noted by \citet{xiong2024llmsexpressuncertaintyempirical}, who attribute it to supervised pretraining that favors confident expressions. After fine-tuning with Rewarding Doubt, the model's confidence scores (shown in \cref{fig:histogramtrained}) span a wider range, including lower values, indicating a more nuanced expression of uncertainty. This shift suggests that fine-tuning mitigates overconfidence and better aligns the model’s confidence with its actual performance.

\begin{table}[htb]
\caption{Comparison of methods on the QAMPARI dataset in the Multiple-Answer setting with 95\% CIs in brackets.}
\centering
\begin{tabular}{lccc}
\toprule
\textbf{Method} & \textbf{ECE} ($\downarrow$) & \textbf{AUROC} ($\uparrow$) & \textbf{Accuracy} ($\uparrow$)\\ 
\midrule
Verbalize & 0.5319 \scriptsize[0.5172,0.5461] & 0.6047 \scriptsize[0.5837,0.6267] & 0.2550 \scriptsize[0.2410,0.2698]\\
Sequence probability & 0.5324 \scriptsize[0.5225,0.5432] & 0.5942 \scriptsize[0.5775,0.6110] & 0.1928 \scriptsize[0.1829,0.2024]\\
Trained probe & 0.1117 \scriptsize[0.0997,0.1262] & 0.6481 \scriptsize[0.6241,0.6726] & 0.2233 \scriptsize[0.2094,0.2384]\\
Rewarding doubt (ours) & \textbf{0.0816} \scriptsize[0.0723,0.0951] & \textbf{0.6947} \scriptsize[0.6776,0.7113] & 0.2480 \scriptsize[0.2348,0.2609]\\
\bottomrule
\end{tabular}
\label{tab:comparison_qampari}
\end{table}

To test the consistency of our method across different models, we  perform an ablation study across diverse LLM architectures and sizes. Specifically, we apply Rewarding Doubt to Qwen-2.5 (3B and 7B) and Gemma-2 (9B) models, in addition to LLaMA-3.1-8B. Table \ref{tab:other_LLM_ablation} reports performance for each model before and after fine-tuning with our method. Despite architectural and pretraining differences, Rewarding Doubt consistently reduces calibration error and improves AUROC across all models, without degrading downstream accuracy.

\begin{table}[tb]
\centering
\caption{Calibration and accuracy of Verbalize vs. Rewarding Doubt across different LLMs with 95\% CIs in brackets.}
\label{tab:other_LLM_ablation}
\begin{tabular}{llccc}
\toprule
\textbf{Model}&\textbf{Method} & \textbf{ECE} ($\downarrow$) & \textbf{AUROC} ($\uparrow$) & \textbf{Accuracy} ($\uparrow$) \\
\midrule
LLaMA-3.1-8B & Verbalize & 0.2771  \scriptsize[0.2689,0.2862] & 0.6766  \scriptsize[0.6667,0.6863] & 0.6662  \scriptsize[0.6577,0.6745] \\
&  Trained probe & \textbf{0.0152}  \scriptsize[0.0118,0.0235] & 0.8495  \scriptsize[0.8420,0.8567] & 0.6231  \scriptsize[0.6143,0.6322] \\
& Rew. Doubt & 0.0256  \scriptsize[0.0209,0.0327] & \textbf{0.8793}  \scriptsize[0.8729,0.8860] & 0.6497  \scriptsize[0.6407,0.6585] \\
\midrule
Qwen-2.5-3B & Verbalize & 0.5330  \scriptsize[0.5252,0.5435] & 0.5981  \scriptsize[0.5927,0.6035] & 0.4185  \scriptsize[0.4085,0.4247] \\
&  Trained probe & \textbf{0.0186}  \scriptsize[0.0134,0.0268] & 0.7975  \scriptsize[0.7880,0.8066] & 0.2540  \scriptsize[0.2463,0.2624] \\
& Rew. Doubt & 0.1483  \scriptsize[0.1415,0.1546] & \textbf{0.9065}  \scriptsize[0.9012,0.9122] & 0.4193  \scriptsize[0.4097,0.4283] \\
\midrule
Qwen-2.5-7B & Verbalize & 0.3619  \scriptsize[0.3530,0.3705] & 0.5818  \scriptsize[0.5762,0.5879] & 0.5239  \scriptsize[0.5148,0.5331] \\
&  Trained probe & \textbf{0.0989}  \scriptsize[0.0920,0.1057] & 0.8737  \scriptsize[0.8676,0.8797] & 0.4793  \scriptsize[0.4696,0.4881] \\
 & Rew. Doubt & 0.1298  \scriptsize[0.1237,0.1368] & \textbf{0.8928}  \scriptsize[0.8866,0.8988] & 0.5283  \scriptsize[0.5193,0.5368] \\
\midrule
Gemma-2-9B & Verbalize & 0.3206  \scriptsize[0.3122,0.3288] & 0.5615  \scriptsize[0.5548,0.5682] & 0.6690  \scriptsize[0.6603,0.6773] \\
&  Trained probe & \textbf{0.0301}  \scriptsize[0.0253,0.0373] & 0.8694  \scriptsize[0.8629,0.8769] & 0.6464  \scriptsize[0.6380,0.6551] \\
& Rew. Doubt & 0.0922  \scriptsize[0.0861,0.0994] & \textbf{0.8649}  \scriptsize[0.8570,0.8725] & 0.6832  \scriptsize[0.6743,0.6918] \\
\bottomrule
\end{tabular}
\end{table}

\paragraph{Stability of Answer Correctness}
Training confidence calibration with our method only targets the uncertainty estimation abilities and does not aim to alter the responses of the model. This is achieved by only rewarding the model on its expressed confidence, while the answer is generated beforehand independently from the model update step. Our results show a stable accuracy for all experiments without notable differences in accuracy between the base model (Verbalize) and the model adapted with Rewarding Doubt, showing that confidence calibration training with Rewarding Doubt does not affect task performance.

\paragraph{Generalization Capabilities} 
To assess the generalization abilities of Rewarding Doubt, we evaluated the model trained on TriviaQA in out-of-domain settings using the CommonsenseQA \citep{talmor-etal-2019-commonsenseqa} and MedQA \citep{jin2020diseasedoespatienthave} datasets. Results are shown in \cref{tab:comparison_methods_commonsense_triviaqatrained}. On MedQA, Rewarding Doubt significantly outperforms Verbalize in both metrics, while on CommonsenseQA, it achieves a comparable ECE, however paired with a much higher AUROC. This discrepancy highlights a limitation of relying solely on ECE for evaluating calibration. ECE does not reflect how well a model discriminates between correct and incorrect predictions across different confidence levels. A model consistently assigning moderate confidence values could appear well-calibrated under ECE, yet fail to offer meaningful distinctions between uncertain and certain cases. AUROC, in contrast, directly measures this discriminative ability. Thus, the substantial improvements in AUROC underscore that Rewarding Doubt produces more useful and actionable confidence estimates. Compared to the Trained Probe, the best-performing baseline, Rewarding Doubt consistently outperforms, showing a stronger ability to generalize to new datasets.

\begin{table}[tb]
\caption{Comparison of generalization results on CommonsenseQA (CsQA) and MedQA, trained on the TriviaQA dataset with 95\% CIs in brackets.}
\centering
\begin{tabular}{l l c c c}
\toprule
 & \textbf{Method} & \textbf{ECE} ($\downarrow$) & \textbf{AUROC} ($\uparrow$) & \textbf{Accuracy} ($\uparrow$)\\
\midrule
\multirow{3}{*}{\rotatebox[origin=c]{90}{CsQA}} 
  & Verbalize 
  & \textbf{0.2820} \scriptsize[0.2206,0.3422] 
  & 0.5425 \scriptsize[0.4740,0.6069] 
  & 0.6860 \scriptsize[0.6277,0.7444] \\
& Trained Probe 
  & 0.4819 \scriptsize[0.4655,0.5130] 
  & 0.5374 \scriptsize[0.5021,0.5708] 
  & 0.7108 \scriptsize[0.6847,0.7355] \\
& Rewarding doubt (ours) 
  & 0.2930 \scriptsize[0.2693,0.3179] 
  & \textbf{0.6385} \scriptsize[0.6065,0.6715] 
  & 0.7163 \scriptsize[0.6918,0.7410] \\
\midrule
\multirow{3}{*}{\rotatebox[origin=c]{90}{MedQA}} 
  & Verbalize 
  & 0.4480 \scriptsize[0.4200,0.4753] 
  & 0.5075 \scriptsize[0.4803,0.5338]
  & 0.5067 \scriptsize[0.4784,0.5350] \\
& Trained Probe 
  & 0.2099 \scriptsize[0.1881,0.2439] 
  & 0.5513 \scriptsize[0.5207,0.5844] 
  & 0.5051 \scriptsize[0.4792,0.5318] \\
& Rewarding doubt (ours) 
  & \textbf{0.1145} \scriptsize[0.0893,0.1408]
  & \textbf{0.6649} \scriptsize[0.6355,0.6954] 
  & 0.5161 \scriptsize[0.4886,0.5420] \\
\bottomrule
\end{tabular}
\label{tab:comparison_methods_commonsense_triviaqatrained}
\end{table}

We also explore generalization across experimental settings in \cref{tab:comparison_qampari_training} by applying a model trained in a Single-Answer setting to a Multiple-Answer task. Although under-performing a model trained specifically for that task, it still outperforms the base model considerably, demonstrating transferability of the learned confidence estimation patterns. This suggests promising applications for improving confidence estimation in more complex or less structured scenarios, such as fact verification and calibration in free-text generation, even when specialized training data is unavailable.

Our current experiments focus on settings where answer quality can be evaluated via exact rule-based metrics, yielding a binary correctness signal, the Rewarding Doubt framework could be extended to work with correctness signals provided by an LLM-as-a-judge system, a reward model trained on human preferences or continuous NLG metrics.

Overall, our experiments show that Rewarding Doubt provides a robust and efficient way to enhance calibration, while generalizing across tasks, and maintaining stable task performance, making it an effective approach for accurate confidence calibration and expression in LLMs. Beyond improvements in calibration, our method also offers practical advantages. While fine-tuning requires an initial training investment, inference remains highly efficient, as only a small, constant number of tokens need to be generated to express confidence. In contrast, zero-shot methods like Chain-of-Thought and Self-Consistency have substantial computational overhead during inference by requiring lengthy reasoning chains or multiple generations. Rewarding Doubt introduces no such overhead, does not rely on an additional model, and directly provides actionable confidence estimates through verbalization directly by the LLM, making it highly suitable for real-world deployment.\looseness=-1

\begin{table}[tb]
\caption{Comparison of the base and fine-tuned model on the Qampari dataset in different settings with 95\% CIs in brackets.}
\centering
\begin{tabular}{l l c c c}
\toprule
\textbf{Training} & \textbf{Evaluation} & \textbf{ECE} ($\downarrow$) & \textbf{AUROC} ($\uparrow$) \\ 
\midrule
Base model & Single fact & 0.5875 \scriptsize[0.5597,0.6151] & 0.5787 \scriptsize[0.5408,0.6125] \\
Single fact & Single fact & \textbf{0.1536} \scriptsize[0.1320,0.1813] & \textbf{0.7240} \scriptsize[0.6889,0.7577] \\
\midrule
Base model & Multi fact & 0.5319 \scriptsize[0.5172,0.5461] & 0.6047 \scriptsize[0.5837,0.6267] \\
Single fact & Multi fact & 0.1777 \scriptsize[0.1679,0.1890] & 0.6617 \scriptsize[0.6468,0.6779] \\
Multi fact & Multi fact & \textbf{0.1061} \scriptsize[0.0935,0.1206] & \textbf{0.7268} \scriptsize[0.7065,0.7468]\\
\bottomrule
\end{tabular}
\label{tab:comparison_qampari_training}
\end{table}

\paragraph{Limitations}
\label{limitations}
Due to computational constraints, we only tested Rewarding Doubt on models with sizes ranging from 3B to 9B parameters. While we expect similar effectiveness on larger models, empirical validation on such scales would be valuable.

\section{Conclusion}
In this work, we propose Rewarding Doubt, a novel approach that enables LLMs to express confidence in their answers more accurately using natural language. We leverage reinforcement learning with a reward function based on the logarithmic scoring rule that incentivizes well-calibrated confidence expressions. Fine-tuning with our method significantly improves the model’s ability to estimate a calibrated confidence, effectively reducing the overconfidence patterns commonly observed in LLMs. This not only enhances the trustworthiness in AI-generated responses but also lays the groundwork for more reliable human-AI collaboration, where models can transparently communicate uncertainty, an essential step toward safer and more accountable AI systems.

\section*{Acknowledgements}
The authors gratefully acknowledge the financial support by the Bavarian Ministry of Economic Affairs, Regional Development and Energy (StMWi) under project ThoraXAI (DIK-2302-0002), and the German Research Foundation (DFG, grant 469106425 - NA 620/51-1).

\section*{Reproducibility Statement}
In order to ensure reproducibility, we describe implementation details of Rewarding Doubt as well as the used baselines in \cref{sec:exp} and \cref{app:baselines}. Further, \cref{sec:prompts} provides the exact prompts used for different experiments. Lastly, we publish our code at \href{https://github.com/pasta99/RewardingDoubt}{\texttt{https://github.com/pasta99/RewardingDoubt}}.

\bibliography{iclr2026_conference}

\begin{thebibliography}{51}
\providecommand{\natexlab}[1]{#1}
\providecommand{\url}[1]{\texttt{#1}}
\expandafter\ifx\csname urlstyle\endcsname\relax
  \providecommand{\doi}[1]{doi: #1}\else
  \providecommand{\doi}{doi: \begingroup \urlstyle{rm}\Url}\fi

\bibitem[Achiam et~al.(2023)Achiam, Adler, Agarwal, Ahmad, Akkaya, Aleman, Almeida, Altenschmidt, Altman, Anadkat, et~al.]{openai2023gpt4}
Josh Achiam, Steven Adler, Sandhini Agarwal, Lama Ahmad, Ilge Akkaya, Florencia~Leoni Aleman, Diogo Almeida, Janko Altenschmidt, Sam Altman, Shyamal Anadkat, et~al.
\newblock Gpt-4 technical report.
\newblock \emph{arXiv}, 2023.

\bibitem[Amouyal et~al.(2023)Amouyal, Wolfson, Rubin, Yoran, Herzig, and Berant]{amouyal2023qampariopendomainquestionanswering}
Samuel~Joseph Amouyal, Tomer Wolfson, Ohad Rubin, Ori Yoran, Jonathan Herzig, and Jonathan Berant.
\newblock Qampari: An open-domain question answering benchmark for questions with many answers from multiple paragraphs, 2023.
\newblock URL \url{https://arxiv.org/abs/2205.12665}.

\bibitem[Azaria \& Mitchell(2023)Azaria and Mitchell]{azaria2023internalstatellmknows}
Amos Azaria and Tom Mitchell.
\newblock The internal state of an llm knows when it's lying, 2023.
\newblock URL \url{https://arxiv.org/abs/2304.13734}.

\bibitem[Bani-Harouni et~al.(2024)Bani-Harouni, Navab, and Keicher]{bani2024magda}
David Bani-Harouni, Nassir Navab, and Matthias Keicher.
\newblock Magda: Multi-agent guideline-driven diagnostic assistance.
\newblock In \emph{International workshop on foundation models for general medical AI}, pp.\  163--172. Springer, 2024.

\bibitem[Carta et~al.(2023)Carta, Romac, Wolf, Lamprier, Sigaud, and Oudeyer]{carta2023grounding}
Thomas Carta, Cl{\'e}ment Romac, Thomas Wolf, Sylvain Lamprier, Olivier Sigaud, and Pierre-Yves Oudeyer.
\newblock Grounding large language models in interactive environments with online reinforcement learning.
\newblock In \emph{International Conference on Machine Learning}, pp.\  3676--3713. PMLR, 2023.

\bibitem[Chen et~al.(2024)Chen, Liang, Wang, Liang, Xiao, Wei, Chen, Hao, Han, and Wang]{chen2024teaching}
Lida Chen, Zujie Liang, Xintao Wang, Jiaqing Liang, Yanghua Xiao, Feng Wei, Jinglei Chen, Zhenghong Hao, Bing Han, and Wei Wang.
\newblock Teaching large language models to express knowledge boundary from their own signals.
\newblock \emph{arXiv preprint arXiv:2406.10881}, 2024.

\bibitem[Cheng et~al.(2024)Cheng, Sun, Liu, Zhang, Yin, Li, Li, He, Chen, and Qiu]{cheng2024can}
Qinyuan Cheng, Tianxiang Sun, Xiangyang Liu, Wenwei Zhang, Zhangyue Yin, Shimin Li, Linyang Li, Zhengfu He, Kai Chen, and Xipeng Qiu.
\newblock Can ai assistants know what they don't know?
\newblock \emph{arXiv preprint arXiv:2401.13275}, 2024.

\bibitem[Chiang et~al.(2023)Chiang, Li, Lin, Sheng, Wu, Zhang, Zheng, Zhuang, Zhuang, Gonzalez, Stoica, and Xing]{vicuna2023}
Wei-Lin Chiang, Zhuohan Li, Zi~Lin, Ying Sheng, Zhanghao Wu, Hao Zhang, Lianmin Zheng, Siyuan Zhuang, Yonghao Zhuang, Joseph~E. Gonzalez, Ion Stoica, and Eric~P. Xing.
\newblock Vicuna: An open-source chatbot impressing gpt-4 with 90\%* chatgpt quality, 2023.

\bibitem[Cosmides \& Tooby(1996)Cosmides and Tooby]{cosmides1996humans}
Leda Cosmides and John Tooby.
\newblock Are humans good intuitive statisticians after all? rethinking some conclusions from the literature on judgment under uncertainty.
\newblock \emph{cognition}, 58\penalty0 (1):\penalty0 1--73, 1996.

\bibitem[Duan et~al.(2024)Duan, Cheng, Wang, Zavalny, Wang, Xu, Kailkhura, and Xu]{duan2024shifting}
Jinhao Duan, Hao Cheng, Shiqi Wang, Alex Zavalny, Chenan Wang, Renjing Xu, Bhavya Kailkhura, and Kaidi Xu.
\newblock Shifting attention to relevance: Towards the predictive uncertainty quantification of free-form large language models.
\newblock In \emph{Proceedings of the 62nd Annual Meeting of the Association for Computational Linguistics (Volume 1: Long Papers)}, pp.\  5050--5063, 2024.

\bibitem[Geng et~al.(2024)Geng, Cai, Wang, Koeppl, Nakov, and Gurevych]{geng2024surveyconfidenceestimationcalibration}
Jiahui Geng, Fengyu Cai, Yuxia Wang, Heinz Koeppl, Preslav Nakov, and Iryna Gurevych.
\newblock A survey of confidence estimation and calibration in large language models, 2024.
\newblock URL \url{https://arxiv.org/abs/2311.08298}.

\bibitem[Grattafiori et~al.(2024)Grattafiori, Dubey, Jauhri, Pandey, Kadian, Al-Dahle, et~al.]{grattafiori2024llama3herdmodels}
Aaron Grattafiori, Abhimanyu Dubey, Abhinav Jauhri, Abhinav Pandey, Abhishek Kadian, Ahmad Al-Dahle, et~al.
\newblock The llama 3 herd of models, 2024.
\newblock URL \url{https://arxiv.org/abs/2407.21783}.

\bibitem[Hadi et~al.(2023)Hadi, Qureshi, Shah, Irfan, Zafar, Shaikh, Akhtar, Wu, Mirjalili, et~al.]{hadi2023survey}
Muhammad~Usman Hadi, Rizwan Qureshi, Abbas Shah, Muhammad Irfan, Anas Zafar, Muhammad~Bilal Shaikh, Naveed Akhtar, Jia Wu, Seyedali Mirjalili, et~al.
\newblock A survey on large language models: Applications, challenges, limitations, and practical usage.
\newblock \emph{Authorea Preprints}, 2023.

\bibitem[Han et~al.(2023)Han, Han, et~al.]{unsloth}
Daniel Han, Michael Han, et~al.
\newblock Unsloth.
\newblock 2023.
\newblock URL \url{http://github.com/unslothai/unsloth}.

\bibitem[Han et~al.(2024)Han, Li, Chen, Shi, Du, Xiao, Liang, and Lin]{han2024enhancing}
Haixia Han, Tingyun Li, Shisong Chen, Jie Shi, Chengyu Du, Yanghua Xiao, Jiaqing Liang, and Xin Lin.
\newblock Enhancing confidence expression in large language models through learning from past experience.
\newblock \emph{arXiv preprint arXiv:2404.10315}, 2024.

\bibitem[Hu et~al.(2022)Hu, yelong shen, Wallis, Allen-Zhu, Li, Wang, Wang, and Chen]{hu2022lora}
Edward~J Hu, yelong shen, Phillip Wallis, Zeyuan Allen-Zhu, Yuanzhi Li, Shean Wang, Lu~Wang, and Weizhu Chen.
\newblock Lo{RA}: Low-rank adaptation of large language models.
\newblock In \emph{International Conference on Learning Representations}, 2022.

\bibitem[Huang et~al.(2023)Huang, Song, Wang, Zhao, Chen, Juefei-Xu, and Ma]{huang2023lookleapexploratorystudy}
Yuheng Huang, Jiayang Song, Zhijie Wang, Shengming Zhao, Huaming Chen, Felix Juefei-Xu, and Lei Ma.
\newblock Look before you leap: An exploratory study of uncertainty measurement for large language models, 2023.
\newblock URL \url{https://arxiv.org/abs/2307.10236}.

\bibitem[Jin et~al.(2020)Jin, Pan, Oufattole, Weng, Fang, and Szolovits]{jin2020diseasedoespatienthave}
Di~Jin, Eileen Pan, Nassim Oufattole, Wei-Hung Weng, Hanyi Fang, and Peter Szolovits.
\newblock What disease does this patient have? a large-scale open domain question answering dataset from medical exams, 2020.
\newblock URL \url{https://arxiv.org/abs/2009.13081}.

\bibitem[Joshi et~al.(2017)Joshi, Choi, Weld, and Zettlemoyer]{JoshiTriviaQA2017}
Mandar Joshi, Eunsol Choi, Daniel~S. Weld, and Luke Zettlemoyer.
\newblock Triviaqa: A large scale distantly supervised challenge dataset for reading comprehension.
\newblock In \emph{Proceedings of the 55th Annual Meeting of the Association for Computational Linguistics}, Vancouver, Canada, July 2017. Association for Computational Linguistics.

\bibitem[Kadavath et~al.(2022)Kadavath, Conerly, Askell, Henighan, Drain, Perez, Schiefer, Hatfield-Dodds, DasSarma, Tran-Johnson, Johnston, El-Showk, Jones, Elhage, Hume, Chen, Bai, Bowman, Fort, Ganguli, Hernandez, Jacobson, Kernion, Kravec, Lovitt, Ndousse, Olsson, Ringer, Amodei, Brown, Clark, Joseph, Mann, McCandlish, Olah, and Kaplan]{kadavath2022languagemodelsmostlyknow}
Saurav Kadavath, Tom Conerly, Amanda Askell, Tom Henighan, Dawn Drain, Ethan Perez, Nicholas Schiefer, Zac Hatfield-Dodds, Nova DasSarma, Eli Tran-Johnson, Scott Johnston, Sheer El-Showk, Andy Jones, Nelson Elhage, Tristan Hume, Anna Chen, Yuntao Bai, Sam Bowman, Stanislav Fort, Deep Ganguli, Danny Hernandez, Josh Jacobson, Jackson Kernion, Shauna Kravec, Liane Lovitt, Kamal Ndousse, Catherine Olsson, Sam Ringer, Dario Amodei, Tom Brown, Jack Clark, Nicholas Joseph, Ben Mann, Sam McCandlish, Chris Olah, and Jared Kaplan.
\newblock Language models (mostly) know what they know, 2022.
\newblock URL \url{https://arxiv.org/abs/2207.05221}.

\bibitem[Kang et~al.(2024)Kang, Wallace, Tomlin, Kumar, and Levine]{kang2024unfamiliar}
Katie Kang, Eric Wallace, Claire Tomlin, Aviral Kumar, and Sergey Levine.
\newblock Unfamiliar finetuning examples control how language models hallucinate.
\newblock \emph{arXiv preprint arXiv:2403.05612}, 2024.

\bibitem[Kapoor et~al.(2024)Kapoor, Gruver, Roberts, Pal, Dooley, Goldblum, and Wilson]{kapoor-etal-2024-calibration}
Sanyam Kapoor, Nate Gruver, Manley Roberts, Arka Pal, Samuel Dooley, Micah Goldblum, and Andrew Wilson.
\newblock Calibration-tuning: Teaching large language models to know what they don`t know.
\newblock In Ra{\'u}l V{\'a}zquez, Hande Celikkanat, Dennis Ulmer, J{\"o}rg Tiedemann, Swabha Swayamdipta, Wilker Aziz, Barbara Plank, Joris Baan, and Marie-Catherine de~Marneffe (eds.), \emph{Proceedings of the 1st Workshop on Uncertainty-Aware NLP (UncertaiNLP 2024)}, pp.\  1--14, St Julians, Malta, March 2024. Association for Computational Linguistics.
\newblock URL \url{https://aclanthology.org/2024.uncertainlp-1.1/}.

\bibitem[Kuhn et~al.(2023)Kuhn, Gal, and Farquhar]{kuhn2023semantic}
Lorenz Kuhn, Yarin Gal, and Sebastian Farquhar.
\newblock Semantic uncertainty: Linguistic invariances for uncertainty estimation in natural language generation.
\newblock \emph{arXiv preprint arXiv:2302.09664}, 2023.

\bibitem[Leng et~al.(2024)Leng, Huang, Zhu, and Huang]{Leng2024TamingOI}
Jixuan Leng, Chengsong Huang, Banghua Zhu, and Jiaxin Huang.
\newblock Taming overconfidence in llms: Reward calibration in rlhf.
\newblock \emph{ArXiv}, abs/2410.09724, 2024.
\newblock URL \url{https://arxiv.org/abs/2410.09724}.

\bibitem[Lin et~al.(2022)Lin, Hilton, and Evans]{lin2022teaching}
Stephanie Lin, Jacob Hilton, and Owain Evans.
\newblock Teaching models to express their uncertainty in words.
\newblock \emph{arXiv preprint arXiv:2205.14334}, 2022.

\bibitem[Manakul et~al.(2023)Manakul, Liusie, and Gales]{manakul2023selfcheckgpt}
Potsawee Manakul, Adian Liusie, and Mark~JF Gales.
\newblock Selfcheckgpt: Zero-resource black-box hallucination detection for generative large language models.
\newblock \emph{arXiv preprint arXiv:2303.08896}, 2023.

\bibitem[Mielke et~al.(2022)Mielke, Szlam, Dinan, and Boureau]{mielke2022reducing}
Sabrina~J Mielke, Arthur Szlam, Emily Dinan, and Y-Lan Boureau.
\newblock Reducing conversational agents’ overconfidence through linguistic calibration.
\newblock \emph{Transactions of the Association for Computational Linguistics}, 10:\penalty0 857--872, 2022.

\bibitem[Moor et~al.(2023)Moor, Banerjee, Abad, Krumholz, Leskovec, Topol, and Rajpurkar]{moor2023foundation}
Michael Moor, Oishi Banerjee, Zahra Shakeri~Hossein Abad, Harlan~M Krumholz, Jure Leskovec, Eric~J Topol, and Pranav Rajpurkar.
\newblock Foundation models for generalist medical artificial intelligence.
\newblock \emph{Nature}, 616\penalty0 (7956):\penalty0 259--265, 2023.

\bibitem[Ouyang et~al.(2022)Ouyang, Wu, Jiang, Almeida, Wainwright, Mishkin, Zhang, Agarwal, Slama, Ray, et~al.]{ouyang2022training}
Long Ouyang, Jeffrey Wu, Xu~Jiang, Diogo Almeida, Carroll Wainwright, Pamela Mishkin, Chong Zhang, Sandhini Agarwal, Katarina Slama, Alex Ray, et~al.
\newblock Training language models to follow instructions with human feedback.
\newblock \emph{Advances in neural information processing systems}, 35:\penalty0 27730--27744, 2022.

\bibitem[Pellegrini et~al.(2025)Pellegrini, {\"O}zsoy, Busam, Wiestler, Navab, and Keicher]{pellegrini2025radialog}
Chantal Pellegrini, Ege {\"O}zsoy, Benjamin Busam, Benedikt Wiestler, Nassir Navab, and Matthias Keicher.
\newblock Radialog: Large vision-language models for x-ray reporting and dialog-driven assistance.
\newblock In \emph{Medical Imaging with Deep Learning}, 2025.

\bibitem[Schulman et~al.(2017)Schulman, Wolski, Dhariwal, Radford, and Klimov]{schulman2017proximal}
John Schulman, Filip Wolski, Prafulla Dhariwal, Alec Radford, and Oleg Klimov.
\newblock Proximal policy optimization algorithms.
\newblock \emph{arXiv preprint arXiv:1707.06347}, 2017.

\bibitem[Shi et~al.(2024)Shi, Li, Ma, Yang, Ma, and Li]{shi2024chopschatcustomerprofile}
Jingzhe Shi, Jialuo Li, Qinwei Ma, Zaiwen Yang, Huan Ma, and Lei Li.
\newblock Chops: Chat with customer profile systems for customer service with llms, 2024.
\newblock URL \url{https://arxiv.org/abs/2404.01343}.

\bibitem[Stengel-Eskin et~al.(2024)Stengel-Eskin, Hase, and Bansal]{stengel2024lacie}
Elias Stengel-Eskin, Peter Hase, and Mohit Bansal.
\newblock Lacie: Listener-aware finetuning for confidence calibration in large language models.
\newblock \emph{arXiv preprint arXiv:2405.21028}, 2024.

\bibitem[Sun et~al.(2024)Sun, Dai, Luo, Chang, and Li]{sun2024lawluomultiagentcollaborativeframework}
Jingyun Sun, Chengxiao Dai, Zhongze Luo, Yangbo Chang, and Yang Li.
\newblock Lawluo: A multi-agent collaborative framework for multi-round chinese legal consultation, 2024.
\newblock URL \url{https://arxiv.org/abs/2407.16252}.

\bibitem[Talmor et~al.(2019)Talmor, Herzig, Lourie, and Berant]{talmor-etal-2019-commonsenseqa}
Alon Talmor, Jonathan Herzig, Nicholas Lourie, and Jonathan Berant.
\newblock {C}ommonsense{QA}: A question answering challenge targeting commonsense knowledge.
\newblock In \emph{Proceedings of the 2019 Conference of the North {A}merican Chapter of the Association for Computational Linguistics: Human Language Technologies, Volume 1 (Long and Short Papers)}, pp.\  4149--4158, Minneapolis, Minnesota, June 2019. Association for Computational Linguistics.
\newblock \doi{10.18653/v1/N19-1421}.
\newblock URL \url{https://aclanthology.org/N19-1421}.

\bibitem[Tao et~al.(2024)Tao, Yao, Ding, Xie, Cao, Sun, Gao, Shen, and Ding]{tao2024trust}
Shuchang Tao, Liuyi Yao, Hanxing Ding, Yuexiang Xie, Qi~Cao, Fei Sun, Jinyang Gao, Huawei Shen, and Bolin Ding.
\newblock When to trust llms: Aligning confidence with response quality.
\newblock \emph{arXiv preprint arXiv:2404.17287}, 2024.

\bibitem[Tian et~al.(2023)Tian, Mitchell, Zhou, Sharma, Rafailov, Yao, Finn, and Manning]{tian2023just}
Katherine Tian, Eric Mitchell, Allan Zhou, Archit Sharma, Rafael Rafailov, Huaxiu Yao, Chelsea Finn, and Christopher~D Manning.
\newblock Just ask for calibration: Strategies for eliciting calibrated confidence scores from language models fine-tuned with human feedback.
\newblock \emph{arXiv preprint arXiv:2305.14975}, 2023.

\bibitem[Touvron et~al.(2023)Touvron, Lavril, Izacard, Martinet, Lachaux, Lacroix, Rozi{\`e}re, Goyal, Hambro, Azhar, et~al.]{touvron2023llama}
Hugo Touvron, Thibaut Lavril, Gautier Izacard, Xavier Martinet, Marie-Anne Lachaux, Timoth{\'e}e Lacroix, Baptiste Rozi{\`e}re, Naman Goyal, Eric Hambro, Faisal Azhar, et~al.
\newblock Llama: Open and efficient foundation language models.
\newblock \emph{arXiv}, 2023.

\bibitem[Tu et~al.(2024)Tu, Azizi, Driess, Schaekermann, Amin, Chang, Carroll, Lau, Tanno, Ktena, et~al.]{tu2024towards}
Tao Tu, Shekoofeh Azizi, Danny Driess, Mike Schaekermann, Mohamed Amin, Pi-Chuan Chang, Andrew Carroll, Charles Lau, Ryutaro Tanno, Ira Ktena, et~al.
\newblock Towards generalist biomedical ai.
\newblock \emph{NEJM AI}, 1\penalty0 (3):\penalty0 AIoa2300138, 2024.

\bibitem[Ulmer et~al.(2024)Ulmer, Gubri, Lee, Yun, and Oh]{ulmer2024calibrating}
Dennis Ulmer, Martin Gubri, Hwaran Lee, Sangdoo Yun, and Seong~Joon Oh.
\newblock Calibrating large language models using their generations only.
\newblock \emph{arXiv preprint arXiv:2403.05973}, 2024.

\bibitem[Wan et~al.(2024)Wan, Huang, Cui, Quan, Bi, and Shi]{wan2024knowledge}
Fanqi Wan, Xinting Huang, Leyang Cui, Xiaojun Quan, Wei Bi, and Shuming Shi.
\newblock Knowledge verification to nip hallucination in the bud.
\newblock \emph{arXiv preprint arXiv:2401.10768}, 2024.

\bibitem[Wang(2024)]{wang2024calibrationdeeplearningsurvey}
Cheng Wang.
\newblock Calibration in deep learning: A survey of the state-of-the-art, 2024.
\newblock URL \url{https://arxiv.org/abs/2308.01222}.

\bibitem[Wang et~al.(2022)Wang, Wei, Schuurmans, Le, Chi, Narang, Chowdhery, and Zhou]{wang2022self}
Xuezhi Wang, Jason Wei, Dale Schuurmans, Quoc Le, Ed~Chi, Sharan Narang, Aakanksha Chowdhery, and Denny Zhou.
\newblock Self-consistency improves chain of thought reasoning in language models.
\newblock \emph{arXiv preprint arXiv:2203.11171}, 2022.

\bibitem[Wei et~al.(2022)Wei, Wang, Schuurmans, Bosma, Xia, Chi, Le, Zhou, et~al.]{wei2022chain}
Jason Wei, Xuezhi Wang, Dale Schuurmans, Maarten Bosma, Fei Xia, Ed~Chi, Quoc~V Le, Denny Zhou, et~al.
\newblock Chain-of-thought prompting elicits reasoning in large language models.
\newblock \emph{Advances in neural information processing systems}, 35:\penalty0 24824--24837, 2022.

\bibitem[Xiong et~al.(2024)Xiong, Hu, Lu, Li, Fu, He, and Hooi]{xiong2024llmsexpressuncertaintyempirical}
Miao Xiong, Zhiyuan Hu, Xinyang Lu, Yifei Li, Jie Fu, Junxian He, and Bryan Hooi.
\newblock Can llms express their uncertainty? an empirical evaluation of confidence elicitation in llms, 2024.
\newblock URL \url{https://arxiv.org/abs/2306.13063}.

\bibitem[Xu et~al.(2024)Xu, Zhu, Zhang, Ma, Fan, Chen, and Yu]{xu2024rejectionimprovesreliabilitytraining}
Hongshen Xu, Zichen Zhu, Situo Zhang, Da~Ma, Shuai Fan, Lu~Chen, and Kai Yu.
\newblock Rejection improves reliability: Training llms to refuse unknown questions using rl from knowledge feedback, 2024.
\newblock URL \url{https://arxiv.org/abs/2403.18349}.

\bibitem[Yang et~al.(2024)Yang, Chern, Qiu, Neubig, and Liu]{yang2024alignment}
Yuqing Yang, Ethan Chern, Xipeng Qiu, Graham Neubig, and Pengfei Liu.
\newblock Alignment for honesty.
\newblock \emph{Advances in Neural Information Processing Systems}, 37:\penalty0 63565--63598, 2024.

\bibitem[Zhang et~al.(2024)Zhang, Diao, Lin, Fung, Lian, Wang, Chen, Ji, and Zhang]{zhang2024r}
Hanning Zhang, Shizhe Diao, Yong Lin, Yi~Fung, Qing Lian, Xingyao Wang, Yangyi Chen, Heng Ji, and Tong Zhang.
\newblock R-tuning: Instructing large language models to say ‘i don’t know’.
\newblock In \emph{Proceedings of the 2024 Conference of the North American Chapter of the Association for Computational Linguistics: Human Language Technologies (Volume 1: Long Papers)}, pp.\  7106--7132, 2024.

\bibitem[Zhang et~al.(2019)Zhang, Kishore, Wu, Weinberger, and Artzi]{zhang2019bertscore}
Tianyi Zhang, Varsha Kishore, Felix Wu, Kilian~Q Weinberger, and Yoav Artzi.
\newblock Bertscore: Evaluating text generation with bert.
\newblock \emph{arXiv preprint arXiv:1904.09675}, 2019.

\bibitem[Zhou et~al.(2023)Zhou, Peng, and Riedl]{zhou2023dialogue}
Wei Zhou, Xiangyu Peng, and Mark Riedl.
\newblock Dialogue shaping: Empowering agents through npc interaction.
\newblock \emph{arXiv preprint arXiv:2307.15833}, 2023.

\bibitem[Zhou et~al.(2025)Zhou, Jin, Shi, and Li]{zhou2025steerconf}
Ziang Zhou, Tianyuan Jin, Jieming Shi, and Qing Li.
\newblock Steerconf: Steering llms for confidence elicitation.
\newblock \emph{arXiv preprint arXiv:2503.02863}, 2025.

\end{thebibliography}
\bibliographystyle{iclr2026_conference}

\newpage
\setcounter{page}{1}
\appendix
\section*{Appendix}
\section{Prompts}
\label{sec:prompts}
For all the question-answering settings, the model is directly prompted to answer a question without a preceding example or context. For our method the model was prompted to answer the question and additionally provide a verbalized confidence. For the other baselines that do not need a verbalized confidence but infer it indirectly, the model is prompted to only give the correct answer. The specifics for multiple-choice are slightly changed but hold mostly the same meaning. The exact prompts for each method can be seen in \cref{tab:triviaqaprompts} for open questions and \cref{tab:mcprompts} for multiple-choice questions. The prompts for each Multi-Answer method can be seen in \cref{tab:multifactprompts}. We decided not to give the model a role like "expert" in the system prompt but keep it neutral, as we observed that the role we give the model affects the verbalized confidence. \\

\begin{table}[h]
    \caption{The prompts for each method for the open short form questions. }
    \centering
    \begin{tabular}{|p{0.16\textwidth}|p{0.75\textwidth}|}
    \hline
        \textbf{Method} & \textbf{Prompt} \\
    \hline
        Verbalize/ Rewarding Doubt & You will get questions. Answer with the correct answer. Additionally provide a confidence between 0, 1, 2, 3, 4, 5, 6, 7, 8, 9, 10, of how sure you are the answer is correct. A value close to 0 means you think there is a high probability that the answer is wrong. The closer the value is to 10, the higher you think is the probability that the answer is correct. The output should have the format 'Answer: <answer>, Confidence: <confidence>' and nothing else. \\
    \hline
         CoT & You will get questions. Read the question, analyze step by step, provide your reasoning steps, answer and your confidence in the response. The confidence should be between 0, 1, 2, 3, 4, 5, 6, 7, 8, 9, 10. A value close to 0 means you think there is a high probability that you could be wrong. The closer the value is to 5, the lower you think is the chance that you could be wrong. The output should have the format 'Thoughts: <reasoning steps>, Answer: <just the answer and nothing else>, Confidence: <confidence>'. \\
    \hline
        Top-K & You will get test questions with possible options. Read the question and enumerate your 5 best guesses and the probability for each that is correct (0\% to 100\%). Just answer with the options and nothing else. The answer should have the format 1. <answer> {<probability to be correct>}\textbackslash n 2. <answer> {<probability to be correct>}\textbackslash n and so on \\
    \hline
         Sequence Probability/ Trained Probe & You will get questions. Answer with the correct answer only and nothing else. \\
    \hline 
        Surrogate Token & \textit{(user)} You will get questions. Answer with the correct answer only and nothing else. \textit{(model)} <response> \textit{(user)} Is the proposed answer: True False \textbackslash n The proposed answer is: \\
    \hline
         Self-Consistency & You will get questions. Read the question, analyze step by step, provide your reasoning steps and answer. The output should have the format 'Thoughts: <reasoning steps>, Answer: <just the answer and nothing else>' \\
    \hline
    \end{tabular}
    \label{tab:triviaqaprompts}
\end{table}

\begin{table}[h]
    \caption{The prompts for each method for multiple choice questions.}
    \centering
    \begin{tabular}{|p{0.16\textwidth}|p{0.75\textwidth}|}
    \hline
        \textbf{Method} & \textbf{Prompt} \\
    \hline
        Verbalize/ Rewarding Doubt & You will get test questions with possible options. Answer with the correct option. Additionally provide a confidence between 0, 1, 2, 3, 4, 5, 6, 7, 8, 9, 10, of how sure you are the answer is correct. A value close to 0 means you think there is a high probability that the answer is wrong. The closer the value is to 10, the higher you think is the probability that the answer is correct. The output should have the format 'Answer: <answer\_index>, Confidence: <confidence>' and nothing else. \\
    \hline
         CoT & You will get test questions with possible options. Read the question, analyze step by step, provide your reasoningsteps, answer and your confidence in the response. The confidence should be between 0, 1, 2, 3, 4, 5, 6, 7, 8, 9, 10. A value close to 0 means you think there is a high probability that you could be wrong. The closer the value is to 5, the lower you think is the chance that you could be wrong. The output should have the format 'Thoughts: <reasoning steps>, Answer: <answer\_index>, Confidence: <confidence>' and nothing else. \\
    \hline
         Sequence Probability/ Trained Probe & You will get test questions with possible options. Answer with the correct option index only and nothing else. \\
    \hline 
        Surrogate Token & \textit{(user)} You will get test questions with possible options. Answer with the correct option index only and nothing else. \textit{(model)} <response> \textit{(user)} Is the proposed answer: True False \textbackslash n The proposed answer is: \\
    \hline
         Self-Consistency & You will get test questions with possible options. Read the question, analyze step by step, provide your reasoningsteps and the correct option index. The output should have the format 'Thoughts: <reasoning steps>, Answer: <answer\_index>' and nothing else. \\
    \hline
    \end{tabular}
    \label{tab:mcprompts}
\end{table}

\begin{table}[h]
    \caption{The prompts for each method for multiple fact questions.}
    \centering
    \begin{tabular}{|p{0.16\textwidth}|p{0.75\textwidth}|}
    \hline
        \textbf{Method} & \textbf{Prompt} \\
    \hline
        Verbalize/ Rewarding Doubt &  Instructions: 
            1. You will get a question with multiple possible answers.
            2. Enumerate all possible answers you know. After each individual answer state your confidence in this answer. The format should be 'Answer: <answer>, Confidence: <confidence> \textbackslash n' for each individual answer. 
            3. The confidence should be an integer number between 0 and 10. 0 means you know for certain the answer is wrong. 10 means you know for certain the answer is correct. 
            4. Do not say anything else. Do not write multiple answers in one answer block.
            5. When asked about dates, answer with the specific year.
            \\
    \hline
         Sequence Probability/ Trained Probe &  Instructions: 
            1. You will get a question with multiple possible answers.
            2. Enumerate all possible answers you know. Write each single answer in this format "Answer: <answer>\textbackslash n" . 
            3. Do not say anything else. Do not write multiple answers in one answer block or any other comments. 
            4. When asked about dates, answer with the specific year.
            5. Do not repeat answers. \\
    \hline
    \end{tabular}
    \label{tab:multifactprompts}
\end{table}

\FloatBarrier

\section{Proof}\label{sec:proofs}
In the following, we prove Proposition \ref{prop_opt} with the reward function 
\begin{equation*}
    R(a, \hat{p}, j) = \begin{cases}
    log(\hat{p}), & \text{if $j(a)$ = 1 (correct)}\\
    log(1 - \hat{p}), & \text{\text{if $j(a)$ = 0 (incorrect)}}
    \end{cases}
\end{equation*}

\begin{proof}
The proof is analogous to the proof that the logarithmic scoring function is a proper scoring function.

Let $f(\hat{p}) = \mathbb{E}[R(a, \hat{p}, j)]$ be the expected reward for all values of $\hat{p}$ and $p^*$:
\[
f(\hat{p}) = p^* \log(\hat{p}) \;+\; (1 - p^*) \log\bigl(1 - \hat{p}\bigr).
\]
Taking the first derivative w.r.t.\ $\hat{p}$:
\[
f'(\hat{p}) 
= \frac{p^*}{\hat{p}} 
\;-\;
\frac{1 - p^*}{1 - \hat{p}}
\quad \]
and setting
\[f'(\hat{p}) = 0 
\implies
p^*(1 - \hat{p}) = \hat{p}(1 - p^*)\
\;\;\Longrightarrow\;\; \hat{p} = p^*
\]
showing the only critical point in $(0,1)$ of $f'$ is at $\hat{p} = p^*$.\\
The second derivative:
\[
f''(\hat{p}) 
= -\frac{p^*}{\hat{p}^2} 
\;-\;
\frac{1 - p^*}{\bigl(1 - \hat{p}\bigr)^2}
\]
is strictly negative for $\hat{p} \in (0,1)$. Hence, $f(\hat{p})$ is concave and has its global maximum at $\hat{p} = p^*$.
\end{proof}

As the logarithm of 0 is undefined, we add a small constant $\epsilon$ in the reward function we use for training:
\begin{equation*}
    R(a, \hat{p}, j) = \begin{cases}
    log(\max(\hat{p}, \epsilon)), & \text{if $j(a)$ = 1 (correct)}\\
    log(\min(1 - \hat{p}, 1 - \epsilon)), & \text{\text{if $j(a)$ = 0 (incorrect)}}
    \end{cases}
\end{equation*}

Through this clipping all confidence predictions between 0 and $\epsilon$, and 1 and $1-\epsilon$, respectively, are rewarded equally. This leads to the model not being able to differentiate between confidence estimations within these ranges. We argue this effect is minor for a sufficiently small $\epsilon$ and can be disregarded in practice.

\section{Clipped Reward Function}
\label{app:clipped_reward}
The clipped reward function as described in \cref{sec:method}, is defined as follows:
\begin{equation}
\label{eq:rewardfunction}
    R(a, \hat{p}, j) = \begin{cases}
    log(\max(\hat{p}, \epsilon)), & \text{if $j(a)$ = 1 (correct)}\\
    log(\min(1 - \hat{p}, 1 - \epsilon)), & \text{\text{if $j(a)$ = 0 (incorrect)}}
    \end{cases}
\end{equation}
where $\epsilon > 0$ is a small positive constant of 0.001 introduced for numerical stability to avoid evaluating the logarithm at zero.

\section{Implementation Details of Baselines}
\label{app:baselines}
For the Sequence Probability, we compute the average probability for each token in the response. In the Self-Consistency method, we let the model explore ten reasoning pathways, and the similarity of each resulting output is evaluated using the BERTScore metric \cite{zhang2019bertscore}. For the trained probe \cite{azaria2023internalstatellmknows}, the original study introduced a custom dataset comprising short statements classified as either true or false. The model's activations in response to these statements were extracted from specific layers, and a multilayer perceptron (MLP) was subsequently trained on these activations to predict the truthfulness of the statements. To ensure a fair comparison, we adapted this methodology to better align with our data by allowing the model to generate answers to training dataset questions and then extracting its activations from the 24th layer for both the statements and their corresponding answers. The labels for each sample were determined following the same evaluation procedure as described in our evaluation framework. For the architecture of the MLP, we employed the same design as \citet{azaria2023internalstatellmknows} and train it for four epochs with a learning rate of 1e-4 until convergence. The exact prompts used for each baseline are provided in \cref{sec:prompts}.

\section{Societal Impact}
\label{sec:impact}
This work introduces a reinforcement learning approach that enables Large Language Models (LLMs) to express calibrated confidence in their factual answers, advancing safe and trustworthy AI deployment. The method improves reliability and uncertainty awareness in LLMs, which is particularly valuable in high-stakes settings such as medicine, law, or customer support, where overconfident errors can have serious consequences. By optimizing a proper scoring rule during training, our method provides a theoretically sound and generalizable mechanism for aligning confidence with factual correctness—supporting human-AI collaboration and informed decision-making. However, expressing numerical confidence may lead users to overly trust AI systems, especially if the model is well-calibrated statistically but still wrong in important individual cases. This risk calls for careful deployment, appropriate user interfaces that contextualize model confidence, and safeguards against overreliance on AI-generated outputs.

\section{Use of Large Language Models}
We employed ChatGPT to enhance the clarity of the manuscript by focusing on grammar corrections, shortening overly complex sentences, and providing alternative wording suggestions. All outputs were manually reviewed before inclusion, and no new technical material, code, results, or figures were generated by the tool.

\end{document}